\documentclass[10pt,twocolumn,letterpaper]{article}

\usepackage[pagenumbers]{cvpr} % To force page numbers, e.g. for an arXiv version

\usepackage{graphicx}
\usepackage{amsmath}
\usepackage{amssymb}
\usepackage{booktabs}
\usepackage{multirow}
\usepackage{float}
\usepackage{caption}
\usepackage[accsupp]{axessibility}

\usepackage[pagebackref,breaklinks,colorlinks]{hyperref}

\usepackage[capitalize]{cleveref}
\crefname{section}{Sec.}{Secs.}
\Crefname{section}{Section}{Sections}
\Crefname{table}{Table}{Tables}
\crefname{table}{Tab.}{Tabs.}

\usepackage{colortbl}
\definecolor{gray}{gray}{.92}
\definecolor{blue}{rgb}{.90,.90,.95}
\definecolor{c1}{rgb}{.98,.92,.92}
\definecolor{c2}{rgb}{.90,.90,.98}

\def\cvprPaperID{1531} % *** Enter the CVPR Paper ID here
\def\confName{CVPR}
\def\confYear{2022}

\begin{document}
\newfloat{figtab}{htb}{fgtb}
\makeatletter
  \newcommand\figcaption{\def\@captype{figure}\caption}
  \newcommand\tabcaption{\def\@captype{table}\caption}
\makeatother

% \title{\LaTeX\ Author Guidelines for \confName~Proceedings}
% \title{Bridging the Present and the Future: Dynamic Context Removal for Action Anticipation}
\title{Learning to Anticipate Future with Dynamic Context Removal}
% \title{Training Efficient Anticipation Model with Dynamic Context Removal} % efficient讲到标题有点虚
\author{
Xinyu Xu\textsuperscript{\rm 1}, Yong-Lu Li\textsuperscript{\rm 1,2}, Cewu Lu\textsuperscript{\rm 1}
\thanks{Cewu Lu is the corresponding author, member of Qing Yuan Research Institute and MoE Key Lab of Artificial Intelligence, AI Institute, Shanghai Jiao Tong University, China and Shanghai Qi Zhi institute.} \\ 
\textsuperscript{\rm 1}Shanghai Jiao Tong University \textsuperscript{2} Hong Kong University of Science and Technology\\
{\tt\small \{xuxinyu2000, yonglu\_li, lucewu\}@sjtu.edu.cn}
}

\maketitle
\begin{abstract}
% 时序推理问题意义，目前现状 - 2 sentences
% 过去方法的问题 - 1 sentence
% 我们的着眼点和insight -- 1
% 具体做法 -- 2-3
% 好处：类人学习，lite，动态根据难度选择课程 -- 1-2
% 实验效果 -- 1
Anticipating future events is an essential feature for intelligent systems and embodied AI. However, compared to the traditional recognition task, the uncertainty of future and reasoning ability requirement make the anticipation task very challenging and far beyond solved. In this filed, previous methods usually care more about the model architecture design or but few attention has been put on how to train an anticipation model with a proper learning policy.
To this end, in this work, we propose a novel training scheme called \textbf{Dynamic Context Removal} (\textbf{DCR}), which dynamically schedules the visibility of observed future in the learning procedure. It follows the human-like curriculum learning process, \ie, gradually removing the event context to increase the anticipation difficulty till satisfying the final anticipation target. Our learning scheme is plug-and-play and easy to integrate any reasoning model including transformer and LSTM, with advantages in both effectiveness and efficiency. 
In extensive experiments, the proposed method achieves state-of-the-art on four widely-used benchmarks.
Our code and models are publicly released at \href{https://github.com/AllenXuuu/DCR}{https://github.com/AllenXuuu/DCR}.
% Our code and models are publicly released at https://github.com/AllenXuuu/DCR.
\end{abstract}

\section{Introduction}
\label{sec:intro}
Anticipating human actions in the near future is a fundamental ability of humans as well as a basic requirement for intelligent systems with reasoning functionality. It serves as a support for many applications like autonomous driving~\cite{rasouli2017they,alvarez2020autonomous} and human-robot interaction~\cite{Robot-Centric,koppula2015anticipating}, where the future prediction of pedestrians and users are essential.

With the rapid evolution of deep learning techniques, the comprehensive understanding and analysis of human action videos attract attention in edging researches. In the traditional recognition field, modern video models~\cite{2stream,R2p1D,tsn,csn,tsm,slowfast,mvt,nonlocal,i3d} leverage spatiotemporal modeling to learn both spatial patterns and temporal logic and achieve significant progress in many video \textit{recognition} tasks~\cite{ek100,kay2017kinetics,goyal2017something}. Besides, there is also a growing interest in action \textit{anticipation}~\cite{ek55,ek100,egteagaze+,50salads,breakfast}. 
% It has similar pattern with action recognition as they both require models to give an category discrimination of the actions, but the observed video segment shifts forward in the anticipation task. 
Similarly, they both expect systems to discriminate the existing actions in videos. Differently, the observed video segment given for systems shifts forward in action anticipation, while action recognition systems have the all the information of videos.
Due to the temporal misalignment between visual observation and target action semantics, action anticipation is a much more challenging task than action recognition. It can hardly be simply treated as classification like video recognition for some reasons. First, the spatial configurations which deep neural networks (DNN) learned in the anticipation task is biased towards the supervision of future action labels, leading to the inaccurate representation of the current visual observation~\cite{rulstm}. Second, the observation has a gap with the start time of action event, which challenges the high-level reasoning ability of model especially in the long-term dense action prediction setting~\cite{actionbanks,Ke_2019_CVPR}.

\begin{figure}[!t]
    \centering
    \includegraphics[width = 0.95\linewidth]{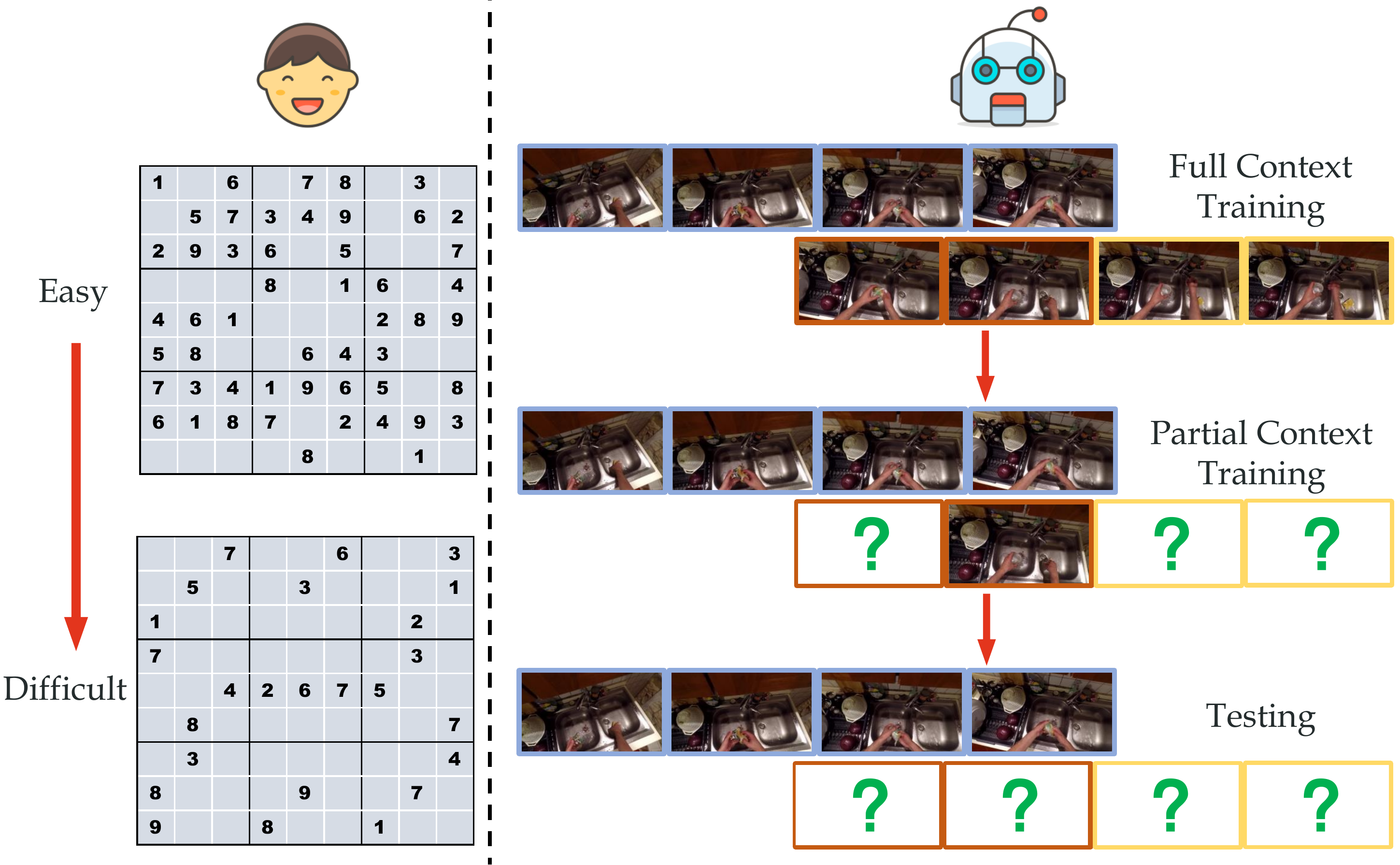}
    \caption{Revisiting learning curriculums in the classical Sudoku game, a kid starts with an easy Sudoku game of more observation (hints) then gets taught a harder level of less observable numbers. This reveals the curriculum learning process of how humans learn to reason in the physical world. In this work, we are inspired by learning Sudoku and build action anticipation model with similar curriculum designs. We leverage extra auxiliary frames in training but dynamically schedule their visibility to gradually strengthen the reasoning ability of model.} % yonglu: good case!
    \label{fig:my_label}
    \vspace{-0.3cm}
\end{figure}

% yonglu: 先讲一段人类思考的方式，有没有心理学或者认知学的证据，比如人类学习sequence的信号时候就是这样先看全部，然后慢慢通过自我的预测尝试和真实反馈逐渐提升自己的预测能力，经验学习blabla；然后再说，按照这一思想，我们采用xxx来学习，类似curriculum learning的思想，但不要说直接用这个，会有a+b的感觉，然后解释逻辑上分几步
To tackle with the action anticipation, previous methods~\cite{rulstm,avt,actionbanks,transaction,red,egoomg,imagineRNN} proposed various neural architectures to focus on learning the temporal logic from past observations, with the intention to apply the past logic in reasoning the future. Though such methods achieve improvements, they still face performance bottlenecks on challenging benchmarks~\cite{ek55,ek100,egteagaze+,50salads,breakfast}. 
We argue that the reason is mainly that they did not learn from the way humans learn.
In this work, we propose a simple but effective perspective for action anticipation. We want the model to learn temporal logic with the auxiliary of the \textit{future snippet} but keep the functionality to reason out the future only given the observation in the past, which meets up the restriction of anticipation problem. To achieve our intention, we propose {\bf D}ynamic {\bf C}ontext {\bf R}emoval {\bf (DCR)} learning scheme, which integrates the motivation of curriculum learning~\cite{bengio2009curriculum} to train with sufficient context auxiliary at first then remove redundant context for better adaptation to a more difficult anticipation task, following the gradual learning process of humans. Fig.~\ref{fig:my_label} gives an intuitive example.

Our training scheme is flexible and can easily advance different temporal reasoning architectures. Here, we mainly choose transformer~\cite{transformer} to implement our paradigm. 
First, in the \textit{full context mode}, we propose the order-aware pre-training to learn video sequential order, which is a generalized method for transformer architecture.
% and we believe it can advance more sequence modeling tasks. 
Next, in the \textit{partial context mode}, we aim to reconstruct frames during action occurrence and dynamically schedule the visibility of auxiliary context. This learning paradigm conforms to how humans learn~\cite{bengio2009curriculum}.
% and indeed builds a better reasoning model. 
Apart from the transformer~\cite{transformer}, we show our training scheme can also improve LSTM~\cite{lstm} based neural architecture.

We conduct experiments and analyses on four widely-used action anticipation benchmarks: EPIC-KITCHENS-100~\cite{ek100}, EPIC-KITCHENS-55~\cite{ek55}, EGTEA GAZE+~\cite{egteagaze+}, 50-Salads~\cite{50salads}. 
Our training strategy turns out to be effective and achieves state-of-the-art on all four benchmarks.
% great performance in the action anticipation problem. 
% Comparing with other methods, we achieve  performances . 
Moreover, we believe the proposed subtractive and adaptive paradigm can pave the way for the other complex and challenging temporal predictive tasks.

Our contribution includes:  
{\bf (1)} We propose a novel learning scheme \textbf{DCR}, which advances the effectiveness and efficiency of practical temporal modeling architectures including transformer~\cite{transformer} and LSTM~\cite{lstm}. 
{\bf (2)} We propose a general order-aware pre-training for transformer architecture to carry out unsupervised pre-training using sequential order as supervision.
{\bf (3)} We achieve state-of-the-arts on four widely-used action anticipation benchmarks.

\section{Related Work}

\noindent\textbf{Action Anticipation} is to predict action in the future by observing video clip with time $\tau_a$ before it occurs. It is required both in third-person~\cite{farha2018what,red,koppula2015anticipating,breakfast,50salads,ant_unlabeled} and egocentric~\cite{ek55,ek100,egteagaze+,rulstm,fhoi,avt,Ke_2019_CVPR,actionbanks,furnari2018leveraging,imagineRNN} scenarios. It has a wide range of applications including intelligent robots~\cite{Robot-Centric,koppula2015anticipating} and wearable devices. It used to have different task formulation such as dense action anticipation~\cite{actionbanks}, but we consider to predict the next action~\cite{ek55,rulstm} in this work. Previous methods proposed various neural architectures including LSTM variants~\cite{jain2016recurrent,farha2018what,red,rulstm,imagineRNN} and attention variants~\cite{actionbanks,transaction,avt}. 
In the early work, Vondrick~\etal~\cite{ant_unlabeled} propose an unsupervised representation learning paradigm to connect the feature of present and future for the anticipation task. 
Li~\etal~\cite{egteagaze+} jointly model action anticipation with human gaze in egocentric videos.
% Zhou~\etal~\cite{zhou2015temporal} jointly model future prediction and sequence ordering as multi-tasks. 
Later, Furnari~\etal~\cite{rulstm} propose a classic RULSTM architecture with modularity attention which 
achieves strong results.
% performs effectively in anticipating future and fusing different modalities. 
Sener~\etal~\cite{actionbanks} attempt to anticipate action with different aggregations on the past. Some other works utilized extra knowledge like next active object~\cite{FURNARI2017401} and hand motion~\cite{egoomg} to anticipation action.
One recent work AVT~\cite{avt} leverages a causal transformer to model action anticipation in the \textit{seq2seq} manner.

\noindent\textbf{Video Sequential Order Modeling} has been exploited in many tasks.
Srivastava~\etal~\cite{srivastava2015unsupervised} propose unsupervised learning techniques to learn generalized representation in video sequence.
Zhou~\etal~\cite{zhou2015temporal} explore two simple tasks of pairwise ordering and future prediction in egocentric videos.
% Vondrick~\etal~\cite{ant_unlabeled} apply unsupervised learning to connect present and future representation to help anticipation.
Kong~\etal~\cite{Kong_2017_CVPR} models sequential context relation to advance the recognition performance on part video observations.
Misral~\etal~\cite{shuffle_and_learn} propose a new perspective of video sequence as to verify whether the order is correct in learning.
In our work, we leverage the permutation invariant property of self-attention and utilize sequential order as extra signals to perform self-supervised learning.

%%%%%%%%%%%%%%%%%%% 改成 training scheme 之后 网络结构可以写弱一点
\noindent\textbf{Vision Transformer} gains much popularity recently, with a trend to exceed the classic convolution architecture in many visual tasks. Transformer~\cite{transformer} family originally raises in the language community, then permeates into the vision domain~\cite{vit} including video related tasks~\cite{mvt,nonlocal,arnab2021vivit}. It can be inserted as attention blocks~\cite{nonlocal,Wu_2019_CVPR} into traditional video models as well as construct pure attention based video recognition architecture~\cite{mvt,arnab2021vivit}. In the filed of video action anticipation, transformer architecture can be directly used in temporal reasoning via causal attention~\cite{avt}.

% \noindent\textbf{Unsupervised Pre-training} is proven to be very effective in many tasks. In the language community, some classic works like Bert~\cite{bert} GPT Series~\cite{gpt2,gpt3} adopt the large-scale pre-training of masked language model and next sentence prediction to achieve dominant improvement for down-stream tasks. For vision tasks, contrastive learning~\cite{moco,simclr} techniques gain much attention in the community, which greatly advances unsupervised representation learning by simply contrasting samples. Video has similar context relations with linguistic sequences and generally shares learning techniques in sequence modeling. Thus, some works~\cite{srivastava2015unsupervised} use the temporal occurrence to supervise the learning or simply connecting the similarity of present and future representation~\cite{ant_unlabeled}. The order of video sequence can be also be used as supervision signal in self-supervised video learning~\cite{zhou2015temporal,shuffle_and_learn}. 

\noindent\textbf{Curriculum Learning} is proposed by Bengio et al.~\cite{bengio2009curriculum}. It is motivated by the learning procedure of human, from easy to hard. It can be implemented via the schedule of category loss weights~\cite{kumar2010self}, data sampling~\cite{li2017multiple} or other difficulty measurement~\cite{zhang2020automatic}. This simple principle works well in many fields including language understanding~\cite{bengio2009curriculum}, transfer learning~\cite{weinshall2018curriculum} and more~\cite{kumar2010self,li2017multiple,zhang2020automatic}. For the language reasoning task, previous work~\cite{cirik2016visualizing} also validates its effectiveness when doing baby-step short-term reasoning first. In our work, the richness of auxiliary context determines the easiness of the task. We schedule the context removal to obey the \textit{easy-to-hard} principle of curriculum learning.

\begin{figure*}[!t]
    \centering
    \includegraphics[width = 1\textwidth]{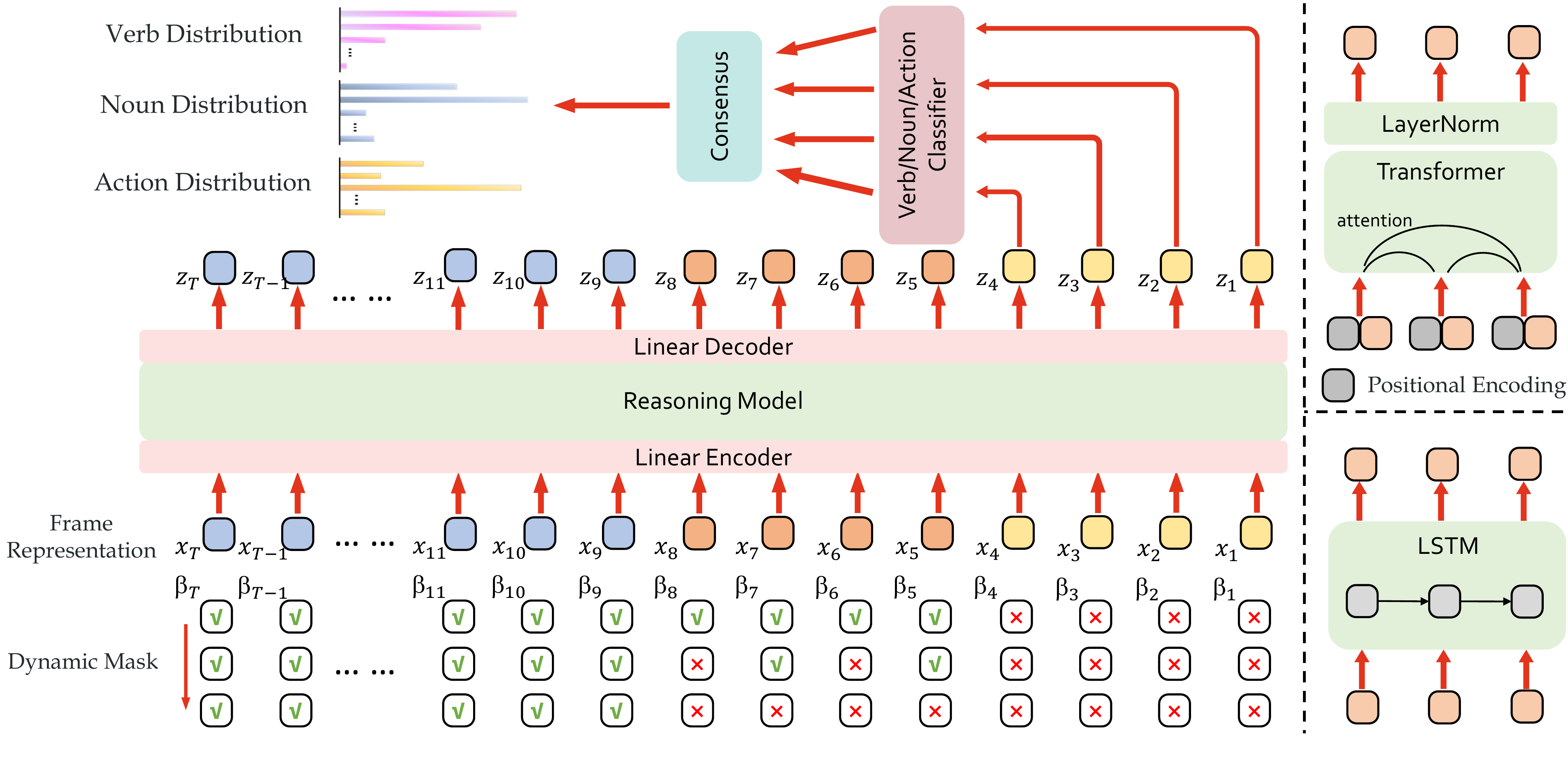}
    \vspace{-0.8cm}
    \caption{An overview of our training scheme. We intend to use past observations to reconstruct frames during action occurrence. The core of our motivation is to schedule the visibility in a curriculum learning manner, with more auxiliary frames at first but dynamically removed as the training goes on. The reconstructed action frames are sent to classifiers and make consensus to obtain final predictions. Our training scheme is flexible and can advance any reasoning models including attention-based transformer and traditional LSTM.}
    \label{fig:arch}
    \vspace{-0.5cm}
\end{figure*}

\begin{figure}[!t]
    \centering
    \includegraphics[width = \linewidth]{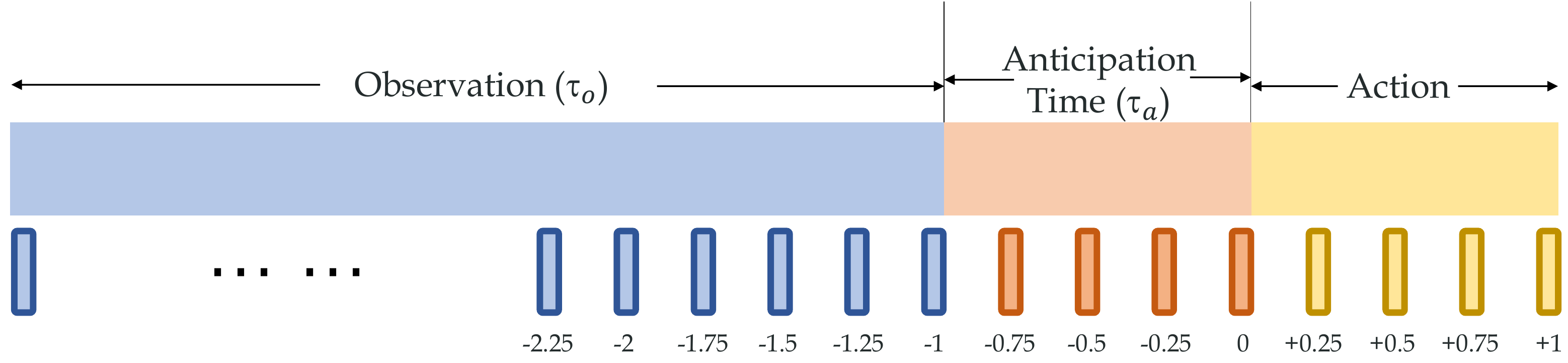}
    \caption{The general setting of the action anticipation task. Blue, red, yellow indicate different duration of past-observed, anticipation, action-occurring respectively.}
    \label{fig:task}
    \vspace{-0.4cm}
\end{figure}

% 我们的目的是根据过去看到的推理出未来的黄色action frame
% 过去的帧包括了蓝色一直可见的红色不可见的
% 红色帧能为黄色帧的推理提供更多帮助, 因此我们用curriculum learning 的方式在学习过程中 利用重建loss作为退火温度的标准 当model学的够好了之后就降温 慢慢不可见红色帧
% 我们的训练方法可以适应较多 推理 网络结构 包括 transf, lstm
% 我们选择transf 开展主要的实验，针对transf 的排序不变性，我们又提出了以permutation 作为监督信号，做full context预训练

% solved: (1) 时间上automatic退火, 用loss做schedule (2) 再加上lstm的实验，可以讲dcr是一个通用的方法
% potential: 空间上如何对红色帧选择扔掉 (attention/loss 其实都没什么道理)

% efficiency: 最好用learnable params 做衡量标准， transf 和 lstm 都要保持比baseline light 但分高 (都用TSN backbone)，然后再喷一下 end2end 训练开销大
% 可能被喷的点: (1) 不知道feature space开始做的讲efficient会不会没什么意思      (2) 采样帧变多了 flops 不好说 

\section{Approach}
\label{sec:approach}
We introduce the core of our work in this section. First, the detailed formulation of anticipation task is introduced in Sec.~\ref{sec:task}. Then, an overview of our method is described in Sec.~\ref{sec:model_arch}. Our motivation is to adapt a well-trained model in the auxiliary context assistance setting into the anticipation setting by dynamic context removal. Thus, we leverage the order-aware pre-training (Sec.~\ref{sec:pretrain}) to learn temporal dynamics for transformer~\cite{transformer} in \textit{full context} mode. In Sec.~\ref{sec:contextrm}, we describe the reconstruction driven curriculum design, which schedules the visibility of context and helps model tackle the anticipation problem gradually. Last, the learning objectives are described in Sec.~\ref{sec:objective}.

\subsection{Task Formulation}
\label{sec:task}
We briefly introduce our action anticipation setting in this section. As illustrated in Figure~\ref{fig:task}, there is a time gap between the observation and the action segment. It is called as anticipation time, expressed as $\tau_a$. We follow previous work~\cite{ek55,ek100,avt,rulstm,imagineRNN,ant_unlabeled,actionbanks,farha2018what} to fix $\tau_a$ = 1s on each benchmark. 
% We build the temporal coordinate whose zero point is the start of action segment. This means we can only observe snippet before -1(included).
Another parameter is $\tau_o$, which denotes the length of observed clip. Usually, $\tau_o$ is not restricted and any choice of $\tau_o$ is permitted. We sample frames at 4 fps following~\cite{rulstm}. We use extra 8 frames ahead to assist training in our framework, but they are strictly not used in the validation and test.

\subsection{Overview}
\label{sec:model_arch}

We present an overview of our learning scheme in Figure~\ref{fig:arch}. Assume we sample $K$ frames for our model, then we start from $K$ pre-extracted frame representations as $x_1,x_2,\cdots,x_K$, in the reverse chronological order. Each frame $x_i$ is assigned with a binary mask $\beta_i\in\{0,1\}$, determining its visibility. The mask is dynamically scheduled in different phase of training (introduced in Sec.~\ref{sec:contextrm}), but we strictly set $\beta_1,\beta_2,\cdots,\beta_8$ = 0 in the test time. We project frame feature into a latent space, where a reasoning model $\mathcal{R}$ performs to reason out the masked frames based on visible information. Then, a linear decoder maps frames back to the original dimension. The goal of our reasoning model is to reconstruct the masked frames and we use $z_1,z_2,\cdots,z_K$ to denote the reconstruction. It is formulated as $z_1,z_2,\cdots,z_K = \mathcal{R}(x_1,\beta_1,x_2,\beta_2,\cdots,x_K,\beta_K)$. The last 4 frames $z_i (1\le i \le 4)$ are frames in the action occurrence and they will be sent to the classifier to give prediction. For EPIC-KITCHENS series~\cite{ek100,ek55} which also require marginalized verb/noun class prediction on their test server, we use verb/noun/action three classifiers on the top, but only apply single action classifier for other datasets. In the test time, predictions on these four frames are averaged to make a consensus~\cite{tsn} as the final result.

Noticeably, our training scheme is flexible and can be used for any reasoning models, including transformer~\cite{transformer}, LSTM~\cite{lstm} \etc. In this paper, we use transformer~\cite{transformer} for most experiments by default, but also give some LSTM~\cite{lstm} based results. A small difference of transformer and LSTM is about tackling masked frames. Mask is more practical for transformer based applications~\cite{bert}, so we directly assign zero value for the input. But for the recurrent LSTM structure, it's more sensitive about latest observation and zero value leads to the smooth prediction of future, thus we copy the masked frame using the latest visible (not masked) one.

% We visualize the detailed architecture in Figure~\ref{fig:arch}. Assume we sample $K$ frames for our model, then we start from $K$ pre-extracted frame representations as $x_1,x_2,\cdots,x_K$, in the reverse chronological order. Each frame $x_i$ is together with a binary mask $\beta_i\in{0,1}$, determining the visibility of it. The mask is dynamically scheduled in different phase of training, but we strictly set $\beta_1,\beta_2,\cdots,\beta_8$ equal 0 in the test time. Frames are multiplied with their masks at first. Next, they are sent into a linear encoder to match the dimensionality of transformer. Then, they get summed with temporal positional encoding before sent into the transformer encoder~\cite{transformer}. We add the post LayerNorm~\cite{ba2016layer} and linear decoder to convert the dimensionality back to the inputs. The reconstructed frames are denoted as $z_1,z_2,\cdots,z_K$ respectively. And we say $z_1,z_2,\cdots,z_K = \mathcal{R}(x_1\beta_1,x_2\beta_2,\cdots,x_K\beta_K)$, where $R$ is our reasoning model. The last 4 frames $z_i (1\le i \le 4)$, which is in the duration of action occurrence, are sent to the classifier $f$ to give prediction. For datasets (EPIC-KITCHENS series) which also require marginalized verb/noun category prediction, we use verb/noun/action three classifiers on the top, but only use action classifiers for other datasets. In the test time, predictions on these frames are averaged to make a consensus~\cite{tsn} as the final result.

\begin{figure}[!t]
    \centering
    \includegraphics[width = .7\linewidth]{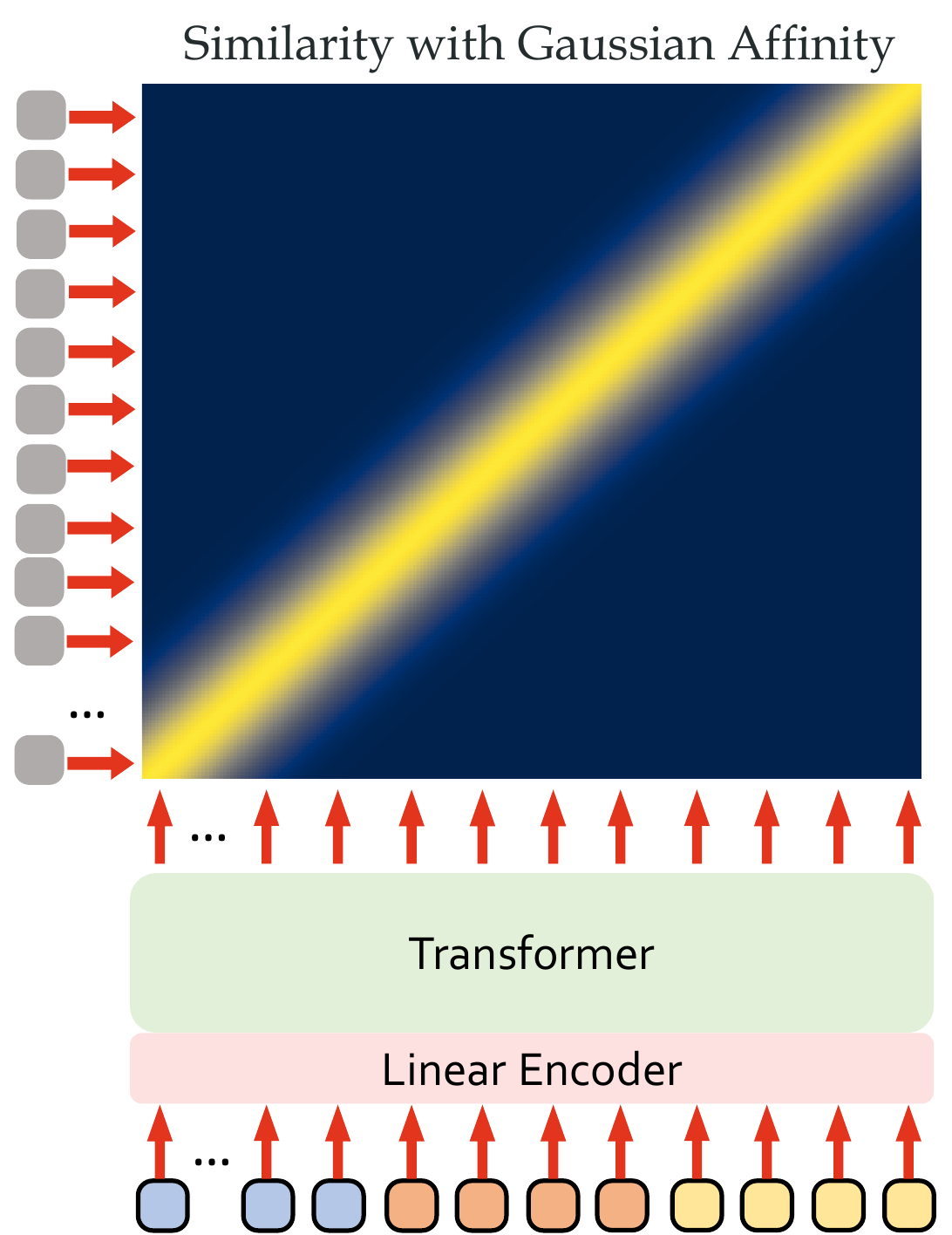}
    \caption{The order-aware pre-training for the permutation-invariant attention. We remove the positional encoding on the input side, but force the model to automatically understand the \textbf{order} of video sequence. It is trained by connecting the frame with its corresponding position to meet the pre-defined similarity.}
    \vspace{-0.3cm}
    \label{fig:pretrain}
\end{figure}

\subsection{Order-Aware Pre-training}
\label{sec:pretrain}
For the transformer based reasoning model, we propose a novel order-aware pre-training to learn the temporal dynamics in the \textit{full context mode}. In this stage, we notice that transformer is a permutation-invariant architecture without explicit positional encoding~\cite{transformer}. Thus, we use temporal position as signal to supervise the training and expect model to automatically recognize the order of input sequence, which implies the understanding of temporal logics among context. Such ordering based temporal modeling turns out to be effective in previous works~\cite{zhou2015temporal}.

We propose our self-supervised pre-training technique called order-aware pre-training. All frames, both the past observation and expanded 8 frames, are used in training. In another word, $\beta_i(1\le i \le K)$ consistently equals 1. Without explicit integration of positional encoding, they are directly sent into the transformer after the linear encoder. Then we compute the cosine similarity between transformer output tokens and the positional encoding, which is followed by Softmax to probability space. 

A pre-defined similarity label is required to supervise the training. The most naive choice is to use a diagonal matrix as similarity to treat it as a separate classification problem. However, time series is continuous, it would be much better to assign soft labels. To this end, we follow~\cite{Hayat_2019_ICCV} to define similarity with Gaussian affinity. The similarity $s_{i,j}$ of positional encoding at time $i$ and the frame feature at time $j$ is measured as
\begin{equation}
\label{eq:sim}
    s_{i,j} = \exp\left(- \frac{(i-j)^2}{\sigma^2} \right)
\end{equation}
where $\sigma$ is the bandwidth of Gaussian and we set $\sigma = 5$ in our experiments. Then the similarity is used to supervise the pre-training. We minimize the cross entropy loss with Gaussian affinity similarity as soft label.

The order-aware pre-training not only learns the relations in the context, but also provides a refinement of positional encoding. It's an aggregated context-agnostic representation on the whole dataset and more suitable for the context-removed scenario in the following training phase. Generally, the technique is motivated by the permutation-invariant property of self-attention and use sequence permutation as self supervised signals. We believe it can advance a wider range of transformer-based sequence modeling tasks, just like the success of masked language model~\cite{bert}.

\subsection{Reconstruction Driven Curriculum Design}
\label{sec:contextrm}

The goal of our expected anticipation model is to reconstruct masked future frames based on the visible context. In this stage, we use \textit{partial context} in training and schedule the context visibility by the reconstruction quality. Here, we inherit the motivation of curriculum learning~\cite{bengio2009curriculum} as the system is given more auxiliary context at the beginning, referred as easy curriculum. We formulate the {\bf easiness} of system as $T_e\in[0,1]$, where $e$ denotes the epoch in training. Specially, we have $T_1=T_2=1$. Then, as the training goes on, we decrease the easiness of the task for our system and present the difficult anticipation task gradually.

In this phase, we consistently mask out the frames during action occurrence as $\beta_i=0(i\le4)$, which are colored yellow in Figure.~\ref{fig:arch}. They are not directly for model input but serve as supervision of the reconstruction. This is also supported by our experiments that direct utilization of action frames harms classifier performance and reasoning model. On the contrary, past observation (blue frames in Figure.~\ref{fig:arch}) are always visible at any time, as $\beta_i=1(i\ge9)$. The median four frames (orange frames in Figure.~\ref{fig:arch}) are the main field for designing different curriculums. They assist past observation to reconstruct anticipated action frames but are dynamically removed determined by the easiness factor $T_e$. For $5 \le i \le 8$ we uniformly sample variable $\rho_i$ in [0,1], as $\rho_i\in U(0,1)$. The frame $x_i$ is visible only when $\rho_i$ is smaller than the easiness factor $T_e$. It also means these frames have a probability $T_e$ to be visible. It's formulated as $\beta_i = \mathbf{1}[T_e > \rho_i] (5 \le i \le 8)$, where $\mathbf{1}[*]$ indicates the truth of statement and returns binary value. Generally, we obtain $\beta$ series in Eq.~\ref{eq:beta}:

\begin{equation}
\label{eq:beta}
    \beta_i = 
    \begin{cases}
        1 & i \ge 9 \\
        \mathbf{1}[T_e > \rho_i] & 5 \le i \le 8\\ 
        0 & i \le 4 \\
    \end{cases}
\end{equation}

Empirically, we design a instance-specific local curriculum scheduling method in this work. Though a global schedule of $T_e$ like linear or exponential may also work well in some scenarios (Sec.~\ref{sec:abl}), we find it's sensitive on hyper-parameters tuning and not very convenient. To this end, we empirically apply an instance-specific easiness schedule. In each iteration, mask $\{\beta_1,\beta_2,\cdots,\beta_K\}$ is generated for a video clip. Assume $k(k\ge5)$ is smallest for $\beta_k = 1$ then $x_1,\cdots,x_{k-1}$ are what we need to anticipate. We use the error of the 1-second future to measure the quality of reconstruction.
\begin{equation}
\label{eq:err}
\begin{aligned}
    Q = ||x_{k-4}-z_{k-4}||_2 \\ 
    s.t. \hspace{.6cm} k = argmin [\beta_k = 1]
\end{aligned}  
\end{equation}

A \textbf{memory bank} is used to store the reconstruction quality for each case. It serves as criterion to define the \textit{easiness} in next epoch. We have $T_1=T_2=1$ at start, but simply schedule easiness $T_e$ using the decline of $Q$ in Eq.~\ref{eq:easiness}, with extra boundaries $\gamma_{min}=0.95,\gamma_{max}=1$ on the decreasing factor. In this case, rapid decline of $Q$ represents a well learnt state of model for this case, thus we decrease easiness faster. The boundaries are used to stabilize easiness scheduling and guarantee the diversity of curriculums in different training stages. 
\begin{equation}
\label{eq:easiness}
    \frac{T_e}{T_{e-1} } = \min\{\max\{\frac{Q_{e-1}}{Q_{e-2}}, \gamma_{min}\},\gamma_{max}\}
\end{equation}

\subsection{Learning Objective}
\label{sec:objective}
We use two objectives to supervise the training process. One is about the predictive result of the next action class $L_{cls}$, while the other is the reconstruction loss of masked frames $L_{rec}$.

% We list the notation of latent variables before introduce losses. Given input frames $x_1,x_2,\cdots,x_K$ with masks $\beta_1,\beta_2,\cdots,\beta_K$, our reasoning model reconstructs frames via $z_1,z_2,\cdots,z_K = \mathcal{R} (x_1 \beta_1,x_2 \beta_2,\cdots,x_K \beta_K)$. $z_i(i \le 4)$ are in the occurrence of action segment, they will be separately sent to the classifier to give prediction $p_{i}^1,p_{i}^2,\cdots,p_{i}^C$, where $C$ is the number of classes. In the test-time, they are averaged to give a consensus of prediction.

\textbf{Prediction loss} is used to supervise the prediction of 4 frames in the action segment. We adopt the cross entropy loss. In addition, we use label smoothing~\cite{szegedy2016rethinking} techniques following~\cite{camporese2021knowledge} and find it works well in our task. This is mainly attributed to the advantages of label smoothing on suppressing overfitting and maintaining uncertainty of future. Assume for $z_i$, action classifier gives prediction $p_{i}^1,p_{i}^2,\cdots,p_{i}^C$, where $C$ is the number of categories. Then, the action prediction loss $L_{cls}^A$ can be formulated in Eq.~\ref{eq:cls_loss}, where $y$ is the ground truth label, $w_y$ is the class loss weight from class distribution and $\epsilon$ is the factor of label smoothing.
\begin{equation}
\label{eq:cls_loss}
    L_{cls}^{A} = \sum_{i=1}^4 - (1-\epsilon) w_y \log(p_{i}^y) - \sum_{j=1} ^ C \frac{\epsilon}{C}  \log(p_{i}^j)
\end{equation}
For datasets which require marginalized verb/noun predictions additionally, we compute verb/noun prediction loss similarly as $L_{cls}^V$, $L_{cls}^N$. The prediction loss is made as $L_{cls} = L_{cls}^V + L_{cls}^N + L_{cls}^A$. For datasets only with an action classifier on the top, we have $L_{cls} = L_{cls}^A$.

\textbf{Reconstruction loss} is to teach our model reason out the masked frames based on the remaining context, just like the role of masked language prediction~\cite{bert}. We expect the output representation of our reasoning model close to the original frame. Thus we simply use mean square error following~\cite{avt} in Eq.~\ref{eq:rec_loss} as feature-level supervision.

\begin{equation}
\label{eq:rec_loss}
    L_{rec} = \sum_{i=1}^K (1- \beta_i) * ||z_i - x_i||_2
\end{equation}

Considering different scales and roles of two losses, we apply a weighted summation to obtain the total loss $L_{total}$:
\begin{equation}
\label{eq:total_loss}
    L_{total} = \lambda_{cls} L_{cls} + \lambda_{rec} L_{rec}
\end{equation}
where $\lambda_*$ are different weights for different loss items.

\begin{figure*}[t] 
    \begin{minipage}[b]{0.37\textwidth} 
        \begin{center}
        \resizebox{\textwidth}{!}{
          \begin{tabular}{ c | ll | ccc | c}
            \toprule
            & Method          & Backbone & Verb  & Noun & Action & \# Params \\
            \midrule
            \rowcolor{c1} & RULSTM~\cite{rulstm}     &    TSN       &  27.5  & 29.0  & 13.3  & 19.7M    \\
            \rowcolor{c1} & AVT~\cite{avt}           &    TSN       &  27.2  & 30.7  & 13.6  & 303.9M   \\
            &    AVT~\cite{avt}                 &    irCSN-152 &  25.5  & 28.1  & 12.8  & 409.6M   \\
            &    AVT~\cite{avt}                 &    ViT*      &  28.7  & 32.3  & 14.9  & 383.8M   \\
            \rowcolor{c1} & DCR (LSTM)               &    TSN       &  27.9  & 28.0  & 14.5  & 14.1M    \\
            &    DCR (LSTM)                     &    TSM       &  28.4  & 28.5  & 15.2  & 20.2M    \\
            \rowcolor{c1} & DCR                      &    TSN       &  31.0  & 31.1  & 14.6  & 78.2M    \\
            
            \multirow{-8}{*}{\rotatebox{90}{RGB}}
            &    DCR                            &    TSM       &  {\bf 32.6} & {\bf 32.7} & {\bf 16.1} & 84.3M  \\
            \midrule\multirow{4}{*}{\rotatebox{90}{Flow}}
            &    RULSTM~\cite{rulstm}       &    TSN      &  19.1  & 16.7  & 7.2   & 19.7M   \\
            &    AVT~\cite{avt}             &    TSN      &  20.9  & 16.9  & 6.6   & 303.9M  \\
            &    DCR (LSTM)                 &    TSN      &  21.6  & 15.3  & 7.8   & 14.1M          \\
            &    DCR                        &    TSN      &  {\bf 25.9} & {\bf 17.6} & {\bf 8.4}  & 78.2M \\
            \midrule\multirow{4}{*}{\rotatebox{90}{Obj}}
            &    RULSTM~\cite{rulstm}       &    FRCNN    &  17.9  & 23.3  & 7.8   & 14.5M   \\
            &    AVT~\cite{avt}             &    FRCNN    &  18.0  & {\bf 24.3}  & 8.7   & 298.8M  \\
            &    DCR (LSTM)                 &    FRCNN    &  16.1  & 19.6  & 7.5   & 10.1M        \\
            &    DCR                        &    FRCNN    &  {\bf 22.2}  & {\bf 24.2}  & {\bf 9.7}   &  74.2M  \\
            \bottomrule
          \end{tabular}
        }
    \end{center}
    \vspace{-0.5cm}
    \tabcaption{Single branch results on EK100~\cite{ek100} validation set. The backbone marked with * denotes end-to-end training.}
    \label{table:result_ek100_val} 
    \end{minipage}
    \hfill
    \begin{minipage}[b]{0.385\textwidth} 
        \begin{center}
        \resizebox{\textwidth}{!}{
          \begin{tabular}{ c | ll | cc | c}
            \toprule
            & Method   &  Backbone &  Top-1  & Top-5 & \# Params\\
            \midrule
            
        \rowcolor{c1}&    RULSTM~\cite{rulstm}           & TSN       & 13.1 &  30.8 & 18.5M             \\
        \rowcolor{c1}&   ActionBanks~\cite{actionbanks} & TSN       & 12.7 &  28.6 & 112.9M            \\
        \rowcolor{c1}&   AVT~\cite{avt}                 & TSN       & 13.1 &  28.1 & 302.6M            \\
            &             AVT~\cite{avt}                 & ViT*      & 12.5 &  30.1 & 382.8M            \\
        \rowcolor{c2}&    AVT~\cite{avt}                 & irCSN-152 & 14.4 &  31.7 & 603.2M            \\
        \rowcolor{c1}&    DCR                            & TSN       & 13.6 &  30.8 & 78.2M             \\ 
        \rowcolor{c2}&    DCR                            & irCSN-152 & 15.1 &  {\bf 34.0} & 82.0M       \\ 
        \multirow{-8}{*}{\rotatebox{90}{RGB}} 
            &             DCR                            & TSM       & {\bf 16.1} &  33.1 & 82.0M       \\ 
            \midrule
            \multirow{3}{*}{\rotatebox{90}{Flow}} 
            &    RULSTM~\cite{rulstm}           &  TSN     & 8.7  &  21.4       & 18.5M     \\
            &    ActionBanks~\cite{actionbanks} &  TSN     & 8.4  &  19.8       & 112.9M     \\
            &    DCR                            &  TSN     & {\bf 8.9}  &  {\bf 22.7} & 78.2M     \\
            \midrule
            \multirow{3}{*}{\rotatebox{90}{Obj}} 
            &    RULSTM~\cite{rulstm}           &   FRCNN  & 10.0 &  29.8    & 13.2M     \\
            &    ActionBanks~\cite{actionbanks} &   FRCNN  & 10.2 &  29.1    & 52.5M     \\
            &    DCR                            &   FRCNN  & {\bf 11.5}  &  {\bf 30.5} &  74.2M   \\
            \bottomrule
          \end{tabular}
        }
    \end{center}
    \vspace{-0.5cm}
    \tabcaption{Single branch results on EK55~\cite{ek55} validation set~\cite{rulstm}. The backbone marked with * denotes end-to-end training.}
    \label{table:result_ek55_val} 
  \end{minipage} 
    \hfill
      \begin{minipage}[b]{0.2\linewidth} 
    \centering 
    \includegraphics[width=1.\linewidth]{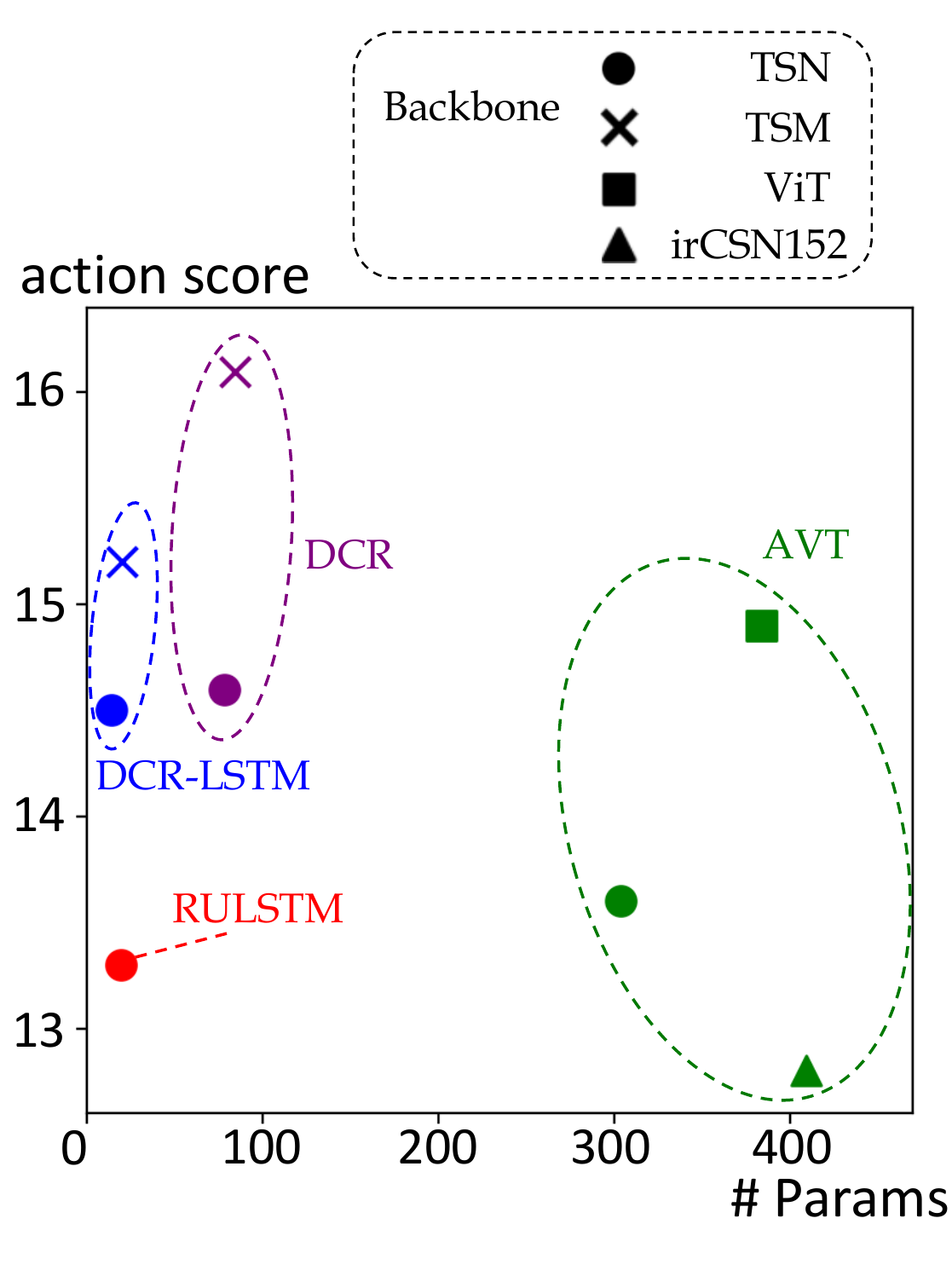} 
    \caption{Score \vs Size.}
    \label{fig:score_and_size} 
  \end{minipage}
  \vspace{-.4cm}
\end{figure*}

\vspace{-0.3cm}

\section{Experiment}

In this section, we conduct comprehensive experiments and analysis on four widely used benchmarks to validate the effectiveness of our method.

\subsection{Datasets and Metrics}
\noindent\textbf{EPIC-KITCHENS-100 (EK100)}~\cite{ek100} is currently the largest dataset to support action anticipation task. It has 700 long videos of 100 hours about egocentric cooking activity. Each action class in EK100 consists of a verb and a noun. Totally, there are 97 verbs and 300 nouns, leading to 4,053 action compositions. There are 89,977 action segments whose labels are aggregated from unique narrations. The dataset splits into train/validation/test sets with the ratio of 75:10:15. The train and validation sets are publicly released but the test set is only able to be queried on the online server. The main metric for evaluation is recall@5, a class aware metric to avoid the long-tail bias of action distribution. Besides, the authors~\cite{ek100} also provide a tail action subset and an unseen 
participants subset to highlight the generalization performance of model.

\noindent\textbf{EPIC-KITCHENS-55 (EK55)}~\cite{ek55} is an earlier version of EK100. As a subset, it contains 432 videos in 55 hours. There are 39,596 action segments, each assigned with a verb and noun class. It includes 125 verbs and 352 nouns in total. We follow the split of~\cite{rulstm,ek55}. The metric for evaluation is Top-1/5 accuracy.
% It should be mentioned that the recall evaluation is different from EPIC-KITCHENS-100, as it only involves many-shot classes provided by the authors~\cite{ek55}.

\noindent\textbf{EGTEA GAZE+ (EG+)}~\cite{egteagaze+} is another egocentric dataset for the joint modeling of action and gaze. We only use its action learning part. It contains 19 verbs, 51 nouns and 106 action compositions. There are 10,325 segments in 86 videos are annotated with action label. We report Top-5 accuracy and class-mean recall@5 over 3 standard official splits provided by the authors~\cite{egteagaze+}.

\noindent\textbf{50-Salads (50S)}~\cite{50salads} is a widely used third-person video dataset about salad preparation. It's a relatively smaller dataset then the previous ones as it only has nearly 0.9K action segments. And differently, its action class can't be marginalized into verb and noun. We follow~\cite{farha2018what,actionbanks,avt} to use the 17-class coarse version of action annotation label. We report Top-1 accuracy over 5 standard official splits provided by the authors~\cite{50salads}.

Following previous works~\cite{ek55,ek100,avt,rulstm,imagineRNN,ant_unlabeled,actionbanks,farha2018what}, we set $\tau_a$ = 1s for all datasets. This setting is also shared for all baselines for a fair comparison.

\subsection{Baseline}
We compare DCR with respect to several competitive approaches including DMR~\cite{ant_unlabeled}, ATSN proposed in~\cite{ek55}, MCE~\cite{furnari2018leveraging}, FHOI~\cite{fhoi}, RULSTM~\cite{rulstm}, ActionBanks~\cite{actionbanks}, ImagineRNN~\cite{imagineRNN}, Ego-OMG~\cite{egoomg}, AVT~\cite{avt} and more. Please refer to supplementary material for more details about baselines.

\subsection{Implementation Detail}

\noindent\textbf{Backbone.} 
We adopt different types of feature (RGB appearance, Optical Flow and Object distribution) from different backbones. The first two modals can be encoded with (1) frame-level spatial model like ViT~\cite{vit} or TSN~\cite{tsn} (2) snippet-level spatialtemporal model like TSM~\cite{tsm} or IG-65M pre-trained irCSN-152~\cite{csn}. Notably, for the spatialtemporal model $\mathcal{B}$ which requires $k$ input frames, we use $x_t = \mathcal{B}\left(I_t,I_{t-1},\cdots,I_{t-(k-1)}\right)$ to encode the feature, where $I_t$ is the raw frame at time $t$. This avoids the involvement of future information. At last, feature of object distribution is represented by the category probability of all objects in the frame and we use the Faster-RCNN(FRCNN)~\cite{faster} detector shared by~\cite{rulstm}. 
Though prior art~\cite{avt} uses a trainable backbone and benefits from the moving regularization of target frame, we consider more about efficiency and choose to freeze backbones in all experiments.

\noindent\textbf{Observation.} We set observation time $\tau_o$ = 10s for EPIC-KITCHENS~\cite{ek100,ek55} series and 50S~\cite{50salads}, but $\tau_o$ = 5s for EG+~\cite{egteagaze+}. Longer observation requirement is mainly because of the larger data scale for EPIC-KITCHENS~\cite{ek100,ek55} and longer action duration in average for 50S~\cite{50salads}.

\noindent\textbf{Head Network.} We use 6-layer, 16-head, 1024-dimensional transformer encoder model~\cite{transformer} optimized by AdamW~\cite{adamw} as the default reasoning architecture but also conduct experiments using  1-layer, 1024-dimensional LSTM~\cite{lstm} optimized by SGD~\cite{adamw} on EK100~\cite{ek100}. We apply learning rate scheduling including 5-epoch warmup and half cosine annealing~\cite{loshchilov2017sgdr} in all experiments. Please refer to supplementary material for details about base learning rate, batch size, loss weight, \etc.

\subsection{Apples-to-Apples Comparison}
We report uni-model performance as well as their trainable parameters on EPIC-KITCHENS series in Tab.~\ref{table:result_ek100_val},\ref{table:result_ek55_val} for a fair comparison. The baseline parameters are recorded from their public checkpoints. Our models have approximative parameters except different dimensionality of input spaces. 

First, we report results on EK100 validation set in Tab.~\ref{table:result_ek100_val}. On the most widely-used RGB-TSN backbone (in red), our LSTM version DCR is slightly lighter than classic RULSTM~\cite{rulstm} while the transformer version is nearly quarter of AVT~\cite{avt} (because of half network width). However, our models consistently perform better, especially for the transformer version, which has 3.8\%/1.0\% performance gains over AVT on verb/action respectively. Additionally, a more effective TSM~\cite{tsm} backbone directly helps DCR to outperform the end-to-end trained AVT by 1.2\% margin at lower expense. We scatter the performance and size of RGB-input models in Fig.~\ref{fig:score_and_size}. Apparently, our methods are in upper left corner, indicating advantages in both effectiveness and efficiency. Besides, on flow and obj modality, our DCR also outperforms previous works. Especially for the flow, we have 5.0\% and 1.2\% performance gains on verb and action respectively.

Next, for results on EK55 validation set in Tab.~\ref{table:result_ek55_val}, our DCR with RGB-TSN backbone also exceeds all baselines (red) in a fair comparison. To our surprise, previous method~\cite{avt} applies 12-layer deep transformer on irCSN-152 backbone to achieve best uni-model performance, but our light model easily outperforms it with 2.3\% gain on Top-5 score (in blue). The stronger TSM backbone further improves top-1 action score by 1.7\% over~\cite{avt}. Besides, our method also achieves competitive results on flow and obj modalities.

Certainly, apples-to-apples comparisons verify the contribution of DCR in training effective anticipation model at lower expense. It clearly paves the way for further research.

\subsection{Comparing to State-Of-The-Art}
%%%%%%%%%%%%%%%%%%%%%%%%%%%%%%%%%%%%%%%%% suppl(3) model ensemble
\noindent{\bf EPIC-KITCHENS}. We late fuse different models to ensemble results on these two benchmarks. Despite previous work may use modality attention~\cite{rulstm} or apply an extra transformer to aggregate multi-modal tokens~\cite{transaction}, our last fusion results still show superiority in Tab.~\ref{table:result_ek100_test},\ref{table:result_ek55_test}. On the validation set, we follow AVT~\cite{avt} to use \textit{rgb+obj} fusion and it outperforms baselines. For example, we have 1.7\% performance gain on the whole EK100 and 3.6\% top-5 action score on EK55. The competitions on the online leaderboard are more challenging. We make ensemble using models trained with \textit{train+val} data. Our method outperforms previous works on most branches, except top-1 score on EK55 S2 test set of unseen participants. This is mainly because the competitive baseline Ego-OMG~\cite{egoomg} adds delicate annotation of hand segmentation and active objects to learn intermediate knowledge representation, which helps anticipation in unseen environments. We argue results on leaderboard rely more on large-scale computation or extra training data. It doesn't matter to validate our effectiveness. For the details of our model ensemble and their weights, please refer to the supplementary material.

\noindent{\bf EGTEA GAZE+}. We use TSN~\cite{tsn} feature on RGB and optical flow modalities following~\cite{rulstm} to conduct this experiment. The final result is the late fusion of two branches in Tab.~\ref{table:result_gaze}. Surprisingly, DCR has 1.5\% and 2.5\% performance gains over all baselines on top-5 accuracy and recall@5 respectively, establishing a new \textit{state-of-the-art}.

\noindent{\bf 50-Salads}. Our training scheme is not limited to egocentric action anticipation, but also advances anticipation results in third-person videos. In this 3-rd view video benchmark, we use same ViT backbone to~\cite{avt} and achieve 3.1\% performance gain on top-1 accuracy score in Tab.~\ref{table:result_50salads}.

\begin{table}[!t]
\begin{center}
\resizebox{0.94 \linewidth}{!}{
  \begin{tabular}{  l | ccc|ccc}
    \toprule
    \multirow{2}{*}{Method} & \multicolumn{3}{c|}{Validation} & \multicolumn{3}{c}{Test} \\
    & Overall & Unseen & Tail & Overall & Unseen & Tail  \\
    \midrule
    RULTSM~\cite{rulstm}                & 14.0  & 14.1  & 11.1  & 11.2  & 9.7   & 7.9  \\
    ActionBanks~\cite{actionbanks}      & 14.7  & 14.5  & 11.8  & 12.6  & 10.5  & 8.9  \\
    TransAction~\cite{transaction}      & 16.6  & 13.8  & 15.5  & 13.4  & 10.1  & 11.9  \\
    AVT~\cite{avt}                      & 15.9  & 11.9  & 14.1  & 16.7  & 12.9  & 13.8  \\
    DCR                                 & {\bf 18.3}  & {\bf 14.7}  & {\bf 15.8}  & {\bf 17.3} & {\bf 14.1} & {\bf 14.3}  \\
    \bottomrule
  \end{tabular}
}
\end{center}
\vspace{-0.6cm}
\caption{Result ensemble on EPIC-KITCHENS-100~\cite{ek100}.}
\vspace{-0.3cm}
\label{table:result_ek100_test}
\end{table}
\begin{table}[t]
\begin{center}
\resizebox{0.95\linewidth}{!}{
  \begin{tabular}{  l | cc|cc | cc}
    \toprule
    \multirow{2}{*}{Method} & \multicolumn{2}{c|}{Validation} & \multicolumn{2}{c|}{Test Seen (S1)} & \multicolumn{2}{c}{Test Unseen (S2)} \\
    & Top-1 &  Top-5 &  Top-1 &  Top-5 &  Top-1 &  Top-5  \\
    \midrule
    ATSN~\cite{ek55}                    & -     & 16.3  &  6.0  & 28.2  & 2.3  & 9.4      \\
    ED~\cite{red}                       & -     & 25.8  &  8.1  & 18.2  & 2.4  & 6.6      \\
    MCE~\cite{furnari2018leveraging}    & -     & 26.1  &  10.8 & 25.3  & 5.6  & 15.7     \\
    RULTSM~\cite{rulstm}                & 15.3  & 35.3  &  14.4 & 33.7  & 8.2  & 21.1     \\
    FHOI~\cite{fhoi}                    & 10.4  & 25.5  &  15.4 & 34.3  & 8.6  & 22.9     \\
    ImagineRNN~\cite{imagineRNN}        & -     & 35.6  &  14.7 & 35.0  & 9.3  & 22.2     \\
    ActionBanks~\cite{actionbanks}      & 15.1  & 35.6  &  16.7 & 36.1  & 10.0  & 23.4    \\
    Ego-OMG~\cite{egoomg}               & {\bf 19.2}  & -     &  16.0 & 34.5  & {\bf 11.8}  & 23.8    \\
    AVT~\cite{avt}                      & 16.6  & 37.6  &  16.8 & 36.5  & 10.4  & 24.3    \\
    DCR                                 & {\bf 19.2}  & {\bf 41.2}  &  {\bf 17.7} & {\bf 38.5}  & 10.9   & {\bf 24.8} \\
    \bottomrule
  \end{tabular}
}
\end{center}
\vspace{-0.6cm}
\caption{Result ensemble on EPIC-KITCHENS-55~\cite{ek55}.}
\vspace{-0.3cm}
\label{table:result_ek55_test}
\end{table}
\begin{table}[t]
\begin{minipage}[b]{0.55\linewidth}
\begin{center}
\resizebox{1\linewidth}{!}{
  \begin{tabular}{ l|cc}
    \toprule
    Method &    Top-5 & c.m. Recall@5 \\
    \midrule
    DMR~\cite{ant_unlabeled}        & 55.7   & 38.1     \\
    ATSN~\cite{ek55}                & 40.5   & 31.6     \\
    MCE~\cite{furnari2018leveraging}& 56.3   & 43.8     \\
    TCN~\cite{bai2018empirical}     & 58.5   & 47.1     \\
    ED~\cite{red}                   & 60.2   & 54.6     \\
    RL~\cite{ma2016learning}        & 62.7   & 52.2     \\
    EL~\cite{jain2016recurrent}     & 63.8   & 55.1     \\
    RULSTM~\cite{rulstm}            & 66.4   & 58.6     \\
    DCR                             & {\bf 67.9}  &  {\bf 61.1} \\
    \bottomrule
  \end{tabular}
}
\end{center}
\vspace{-0.5cm}
\caption{Results on EG+~\cite{egteagaze+}.}
% \vspace{0.5cm}

\label{table:result_gaze}
\end{minipage}
\hfill
\begin{minipage}[b]{0.44\linewidth}
\begin{center}
\resizebox{0.9\linewidth}{!}{
  \begin{tabular}{ l|c}
    \toprule
    Method &    Top-1\\
    \midrule
    DMR~\cite{ant_unlabeled}             & 6.2  \\
    RNN~\cite{farha2018what}             & 30.1 \\
    CNN~\cite{farha2018what}             & 29.8 \\
    ActionBanks~\cite{actionbanks}       & 40.7  \\
    AVT~\cite{avt}                       & 48.0   \\
    DCR                                  & {\bf 51.1}     \\
    \bottomrule
  \end{tabular}
}
\end{center}
\vspace{-0.5cm}
\caption{Results on 50S~\cite{50salads}.}
% \vspace{0.5cm}
\label{table:result_50salads}
\end{minipage}

\vspace{-0.6cm}
\end{table}
\begin{figure*}[t] 
    \begin{minipage}[b]{0.25\textwidth} 
        \begin{center}
        \resizebox{.95\linewidth}{!}{
          \begin{tabular}{ l|c|c|c}
            \toprule
            & \multicolumn{2}{c|}{EK100~\cite{ek100}} & EG+~\cite{ek55} \\
            & TSM & TSN & TSN \\ 
            \midrule
            \textbf{DCR}                            &  {\bf 16.1} &  {\bf 14.6} & {\bf 64.5} \\
            \midrule
            classification                          &  13.7       &   12.7   &  58.5  \\    
            \midrule
            \textit{w.o.}  pre-training             &  15.5       &   14.3   &  62.1  \\
            \midrule
            $T_e$ = 1                               &  6.5        &   4.5    &  40.1  \\
            $T_e$ = 0                               &  15.2       &   13.8   &  62.9  \\
            linear $ T_e$                           &  15.0       &   13.9   &  64.0  \\
            exponential $T_e$                       &  15.6       &   14.2   &  64.2  \\
            \midrule
            \textit{w.o.} $L_{rec}$                 &  13.5       &   12.6   &  56.0  \\
            \textit{w.o.} label smooth              &  14.8       &   13.3   &  62.3  \\
            \bottomrule
          \end{tabular}
        }
        \end{center}
    \vspace{-0.5cm}
    \tabcaption{Ablation study.}
    \label{table:abl} 
    \end{minipage}
    \hfill
    \begin{minipage}[b]{0.24\textwidth} 
            \centering
            \includegraphics[width = \linewidth]{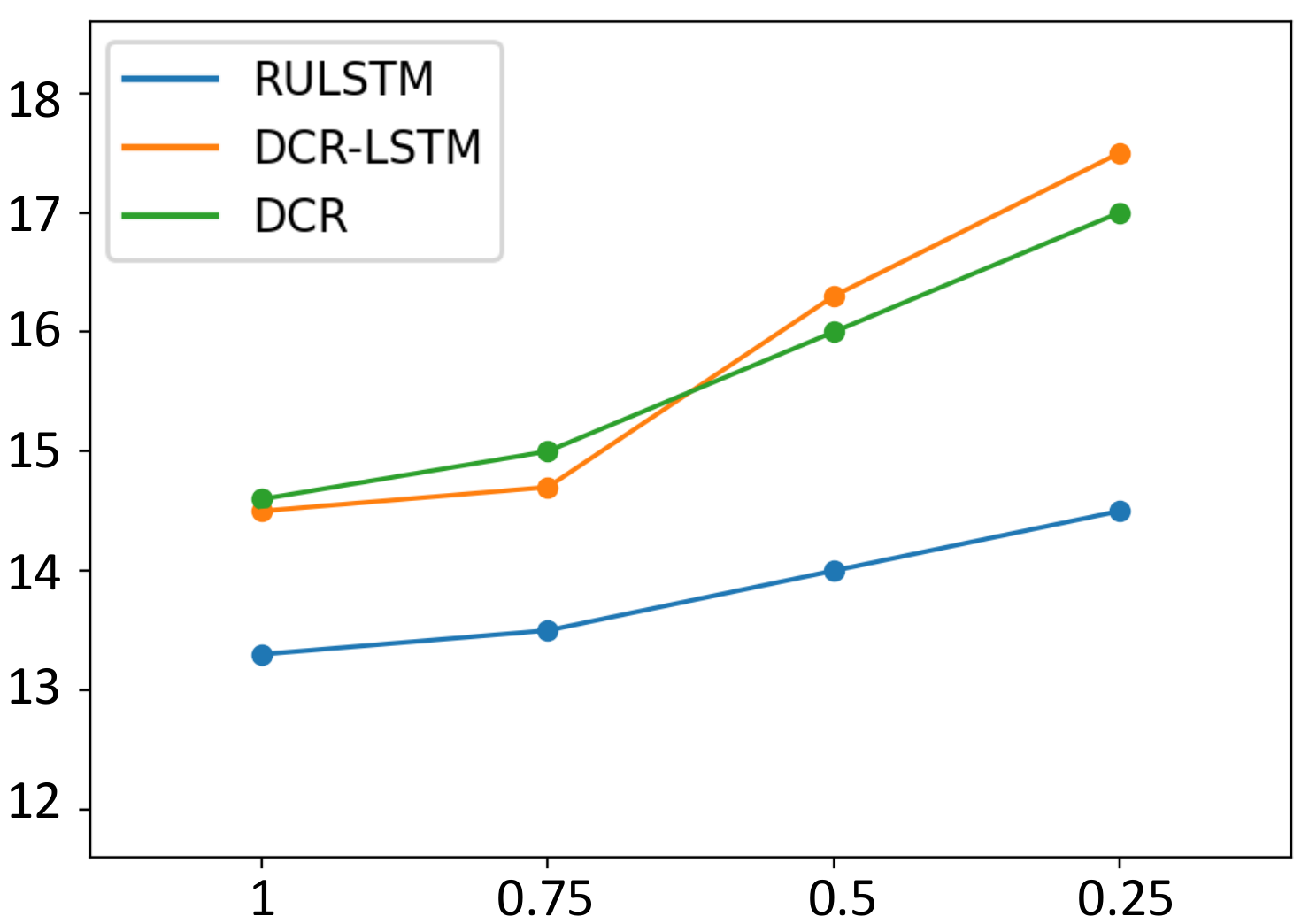}
            \vspace{-.5cm}
            \caption{Effect of richer context via decreasing $\tau_a$.}
            \label{fig:tau_a}
    \end{minipage} 
    \hfill
    \begin{minipage}[b]{0.48\linewidth} 
        \centering
        \includegraphics[width =.95 \linewidth]{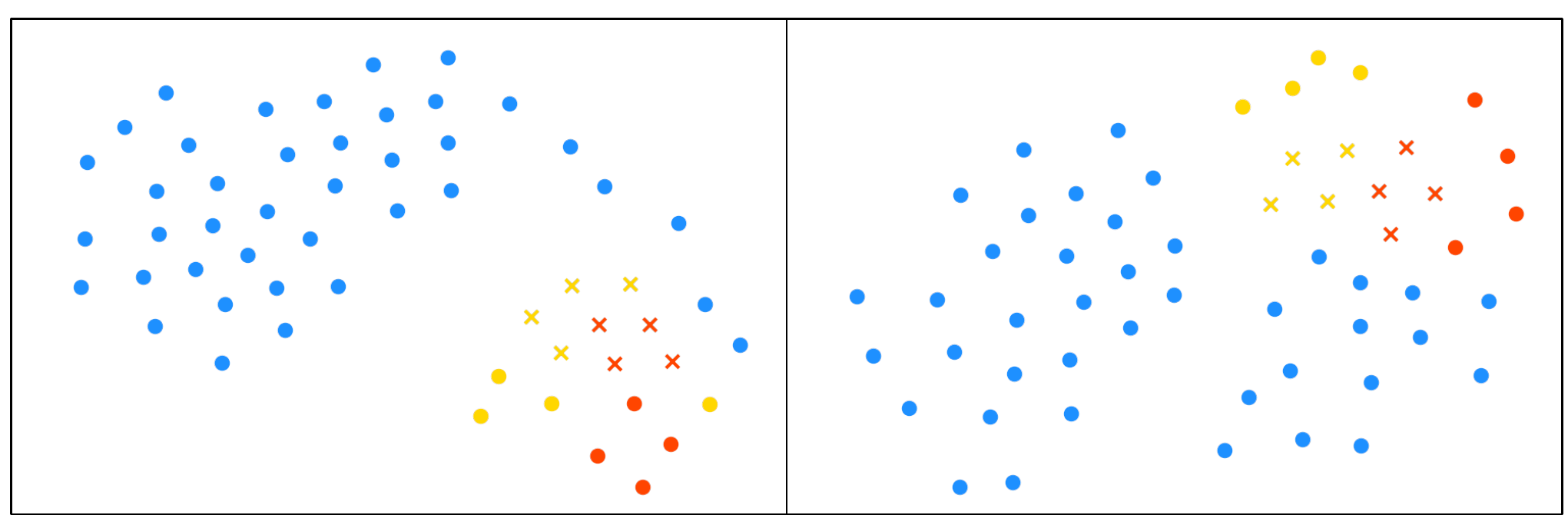}
        \vspace{-.3cm}
        \caption{Qualitative cases of frame reconstruction. Blue dots are visible frames of past. Crosses are the predicted future representation, much closer to the actual future (yellow and red dots).}
        \label{fig:reconstruction}
    \end{minipage}
  \vspace{-.4cm}
\end{figure*}

\subsection{Ablation Study}
\label{sec:abl}
We conduct ablation study to verify the effects of our method on transformer~\cite{transformer} in Tab.~\ref{table:abl}, but leave LSTM~\cite{lstm}-based results in the supplementary material. We report on EK100~\cite{ek100} and EG+~\cite{egteagaze+} with RGB inputs. (1) First, we compare a classification baseline by removing everything used in anticipation task. Each branch has a large performance drop, indicating basic classification technique is not suitable for direct anticipation. (2) Second, we consider model without order-aware pre-train. Their performance can't achieve best, especially 2.4\% drop on EG+. (3) Third, we consider different easiness schedules. If we always train under $T_e$=1, it turns out that training and testing tasks have a large gap and model can't transfer well. If we always train without using future context as $T_e$=0, then the model gets trapped in local optimum and performs not very well. We consider different \textit{global} schedules of $T_e$, like linearly decreases from 1 to 0 or exponentially multiplies $\gamma=0.95$ after each epoch. These methods also bring advance in training the model, but are empirically worse than our \textit{local} schedule proposition. (4) Last, we validate effects of loss components. Our model turns to have largest performance drop without $L_{rec}$, even worse than the classification. This is because different context complicates classification without feature-level supervision. Moreover, without label smoothing, we observe quick loss decreasing in training and worse performance due to overfitting.

\begin{figure}[t]
  \centering
  \includegraphics[width = .95\linewidth]{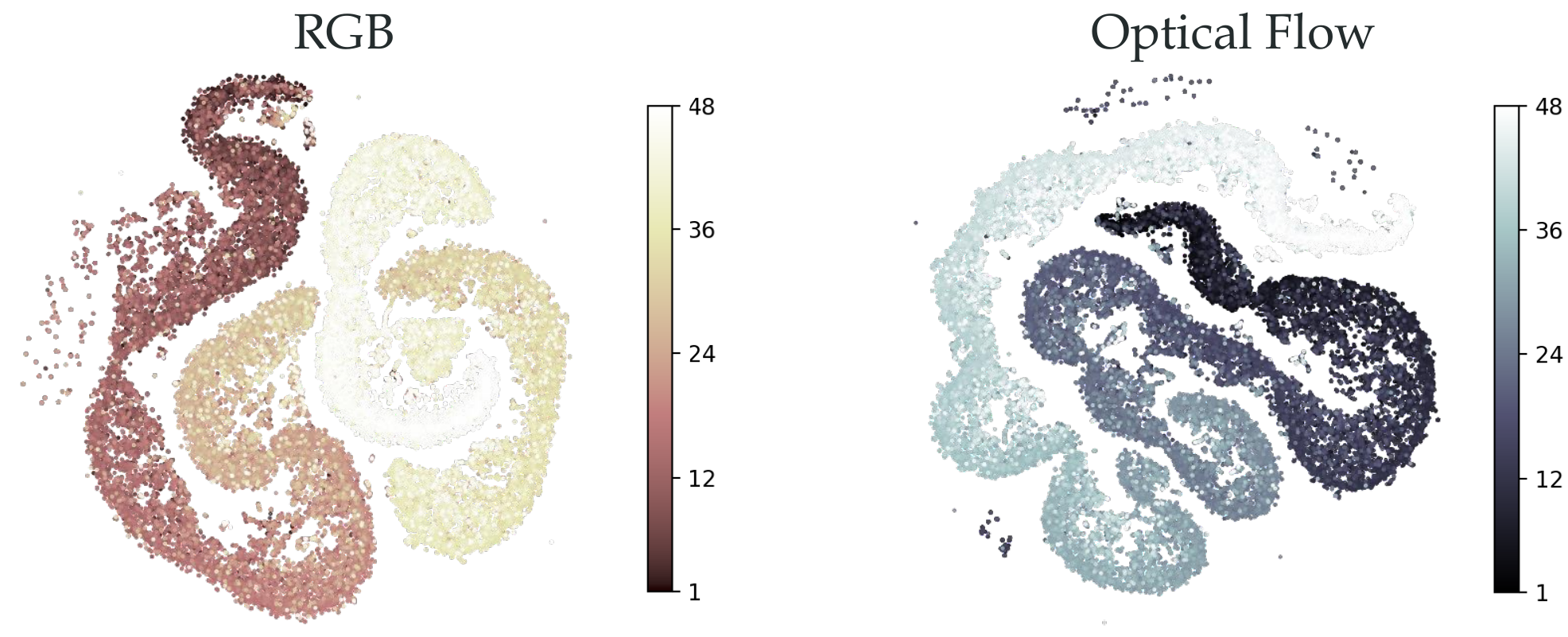}
  \caption{Effect of order-aware pre-training. We sample 1,000 segments from EK100 validation set and color the order-aware tokens from the pre-training models according to their temporal position. Our model can clearly embed the temporal dynamics within its learned manifold.}
  \label{fig:res_order_pretrain}
  \vspace{-0.6cm}
\end{figure}

\subsection{Qualitative Results}
We give qualitative results to better characterize the reasoning ability of our method. 

First, we show what the model learns in the order-aware pre-training phase. We sample 1,000 segments from EK100 validation set, and extract output tokens from pre-trained order-aware transformer. In Fig.~\ref{fig:res_order_pretrain}, we use t-SNE~\cite{tsne} to embed them into a 2D space and color frames according to their temporal position. It's an interesting visualization that indicating our pre-trained models have learnt video dynamics in the latent manifold, with a more comprehensive understanding of temporal logics in video.

Second, effect of auxiliary context shows in Fig.~\ref{fig:tau_a}. We reduce anticipation time $\tau_a$ and test model performance. Thanks to our curriculum learning approach, our model retain stronger predictive ability in easy task. Compare to~\cite{rulstm}, our models have larger boost on more context. LSTM version DCR is more sensitive about latest context and even outperforms the transformer.

Last, we show qualitative cases of frame reconstruction. All frames and the model reconstruction are embedded via t-SNE~\cite{tsne} and visualized in Fig.~\ref{fig:reconstruction}. Blue ones are observed in anticipation task, while the yellow and red ones are the future. The reconstructed frames marked as cross are more close to the cluster of future frames.

\section{Discussion}
\noindent\textbf{Limitations}. We present an intuitive and empirical approach in video action anticipation, \eg the local easiness scheduling. More finer-grained analysis in the future work is required to verify the curriculum learning approach.

\noindent\textbf{Potential Negative Societal Impact}. We train anticipation model on human-annotated datasets, which may import bias from human-defined labels. Due to the uncertainty of future and the potential bias, current anticipation model can hardly make robust prediction about the future action. But a possible solution for debiasing is to utilize unsupervised learning technique on a larger scale of data. On the usage of our method, video action anticipation technique is generally harmless except some malicious use for bad-intended events prediction. Thus, we encourage a proper use of technology that benefits mankind. 

\section{Conclusion}

In this paper, we propose a novel strategy \textbf{DCR} on how to train an anticipation model. It follows the intuitive learning process of humans and flexibly advances any reasoning models in effectiveness and efficiency. In extensive experiments, we establish new \textit{state-of-the-art} on four widely used benchmarks. We believe video action anticipation is an important problem for artificial intelligence, which supports many future applications. We are taking a great step in this field. In addition, our method is not limited to anticipation problem, but also has the potential to boost many other temporal predictive tasks. We hope our simple motivation of dynamic context removal can inspire more future works.

\section*{Acknowledgement}
We appreciate the support from National Natural Science Foundation of China (No.72192821, 72192820), Shanghai Municipal Science and Technology Major Project (2021SHZDZX0102), Shanghai Qi Zhi Institute, and SHEITC (2018-RGZN-02046).

{
    \small
    \bibliographystyle{ieee_fullname}
    \bibliography{egbib}
}

\clearpage

\appendix

\twocolumn[{%
\renewcommand\twocolumn[1][]{#1}%
\maketitle
\begin{center}
    \centering
    {\Large\textbf{Supplementary Material for Learning to Anticipate Future with Dynamic }}\\
    \vspace{2mm}
    {\Large\textbf{Context Removal}}\\
    \vspace{6mm}
    {\large Xinyu Xu\textsuperscript{\rm 1}, Yong-Lu Li\textsuperscript{\rm 1,2}, Cewu Lu\textsuperscript{\rm 1}}\\ 
    \vspace{2mm}
    {\large \textsuperscript{\rm 1}Shanghai Jiao Tong University \textsuperscript{2} Hong Kong University of Science and Technology}\\
    {\tt\small \{xuxinyu2000, yonglu\_li, lucewu\}@sjtu.edu.cn}

    \vspace{8mm}
\end{center}
}]

\section{License}

\subsection{Dataset}
We use four datasets in this work. They are EPIC-KITCHENS-100~\cite{ek100}, EPIC-KITCHENS-55~\cite{ek55}, EGTEA GAZE+~\cite{egteagaze+} and 50-Salads~\cite{50salads}.

EPIC-KITCHENS-100~\cite{ek100} and EPIC-KITCHENS-55~\cite{ek55} are copyright by the same team and published under the Creative Commons Attribution-NonCommercial 4.0 International License~\cite{eklicense}. We download data from its website~\cite{ekwebsite}.

EGTEA GAZE+~\cite{egteagaze+} is publicly available, and no license is specified. We download data from its website~\cite{eg+web}.

50-Salads~\cite{50salads} is licensed under a Creative Commons Attribution-NonCommercial-ShareAlike 4.0 International License~\cite{50saladslicense}. We download data from its website~\cite{50sweb}.

\subsection{Prior Work}
We sincerely thank prior work RULSTM~\cite{rulstm} and AVT~\cite{avt} for their generous release of checkpoints and pre-extracted feature which greatly helps our experiments.

RULSTM~\cite{rulstm} is publicly released in~\cite{rulstmweb}, and no license is specified.

AVT~\cite{avt} is publicly released in~\cite{avtweb}, and licensed under the Apache License 2.0~\cite{avtlicense}.

\section{Baseline}

\noindent\textbf{Deep Multimodal Regressor (DMR)}~\cite{ant_unlabeled} applies an unsupervised training scheme to minimize the representation gap between current observation and the multimodal future via deep regression network. Then the anticipative feature is sent to classifier to give results.

\noindent\textbf{Anticipation TSN (ATSN)}~\cite{ek55} is a variant of TSN~\cite{tsn} that has same model architecture but different input segment. The observed segment is sent to TSN architecture for a simple classification using future action label.

\noindent\textbf{Verb-Noun Marginal Cross Entropy Loss (MCE)}~\cite{furnari2018leveraging} is an effective loss function to boost the anticipation performance of ATSN. It focuses on predicting verb-noun composed action label, but still follows marginal constraints.

\noindent\textbf{Forecasting HOI (FHOI)}~\cite{fhoi} adopts intentional hand movement and jointly predicts the egocentric hand motion, interaction hotspots and future action.

\noindent\textbf{RULSTM}~\cite{rulstm} is the winner of EK55 2019 anticipation challenge~\cite{ek55}. It utilizes one rolling LSTM to summarize the past and another unrolling LSTM to anticipate future. Modality attention mechanism (MATT) is proposed to make multi-modal prediction.

\noindent\textbf{ImagineRNN}~\cite{imagineRNN} attempts to anticipate future by imagination. It narrows the gap of observation and action execution with an imagined intermediate and further improves performance by the residual anticipation.

\noindent\textbf{ActionBanks}~\cite{actionbanks} is the winner of EK55 2020 anticipation challenge~\cite{ek55}. It leverages different levels of past aggregation representation via attention mechanism to improve the anticipation performance.

\noindent\textbf{Ego-OMG}~\cite{egoomg} annotates extra information about hand segmentation and (next) active objects to serve as intermediate knowledge in anticipation. It is encoded by graph network and LSTM then ensembled with additional CSN~\cite{csn} branch to give the prediction.

\noindent\textbf{Anticipative Video Transformer (AVT)}~\cite{avt} is a recent work as well as the winner of EK100 2021 anticipation challenge~\cite{ek100}. It proposes an end-to-end transformer~\cite{transformer} based architecture with causal attention on the head to anticipate future in the \textit{seq2seq} manner.

\section{Model Ensemble}

We simply last fuse results from different models to give prediction on EPIC-KITCHENS~\cite{ek100,ek55} series, and it also performs more complex fusion methods~\cite{rulstm,transaction}. Noticeably, We only use the transformer~\cite{transformer} version DCR in fusion. For EK100 validation set, we fuse TSM-DCR, TSN-DCR, FRCNN-DCR with weight 1:1:1. For EK55 validation set, we fuse TSM-DCR, irCSN152-DCR, TSN-DCR, FRCNN-DCR with weight 1:1:1:1. The submissions to online test server is more challenging since the huge computation cost and additional training data of competitive baselines. Thus, we leverage the public model zoo from prior work AVT~\cite{avt}, which makes seven model ensemble and achieves \textit{state-of-the-art}. On EK100 test set, we fuse results of TSM-DCR, TSN-DCR, AVT with weight 1:0.5:1. On EK55, we fuse TSM-DCR, irCSN152-DCR, AVT with weight 1:1:1 for test set S1, while 0.5:1.5:1.5 for test set S2. Results are listed in main text.

\begin{table}[!t]
\begin{center}
\resizebox{1\linewidth}{!}{
  \begin{tabular}{lcccccc}
    \toprule
                         & $\epsilon$ & $\lambda_{cls}$ & $\lambda_{rec}$ & learning rate & batch size  & epoch \\
    \midrule
    EK100~\cite{ek100}   &        0.2 &             0.5 &               1 &          1e-4 &        128  &    100\\
    EK55~\cite{ek55}     &        0.4 &             1   &               1 &          1e-4 &        128  &    100\\
    EG+~\cite{egteagaze+}&        0.4 &             0.5 &               1 &          5e-5 &        512  &    50 \\
    50S~\cite{50salads}  &        0.5 &             0.5 &               2 &          5e-5 &        64   &    50 \\
    \bottomrule
  \end{tabular}
}
\end{center}
\vspace{-0.5cm}
\caption{Hyper-parameters for DCR training details with transformer~\cite{transformer} head.}
\label{table:hyper}
\end{table}

\section{Additional Training Detail}

We present additional training details as a supplement to Sec. 4.3, main text. The default transformer architecture starts with order-aware pre-training. In this phase, for all datasets, we set batch size 512 and optimize the network for 50 epoch with base learning rate is 1e-4. Then, the next stage is reconstructing future with dynamic context. We customize hyper-parameters for different datasets in Tab.~\ref{table:hyper}. We can conclude some interesting empirical results in hyper-parameter tuning, \eg small dataset suffers more from future uncertainty and increasing label smoothing~\cite{szegedy2016rethinking} level is beneficial. We conduct LSTM~\cite{lstm} experiments on EK100~\cite{ek100}. It's not applied with pre-training but directly starts with the second stage. It is optimized with base learning rate 1e-2 for 100 epochs. We set batch size = 512, $\lambda_{cls}$=1, $\lambda_{rec}$=1, smoothing factor $\epsilon$=0.2.

% We give most of training details in Sec. 4.3, main text. Here we present additional dataset-customized hyper-parameter settings for DCR training details with transformer~\cite{transformer} head in Tab.~\ref{table:hyper}. We can conclude some interesting empirical results in hyper-parameter tuning, \eg small dataset suffers more from future uncertainty and increasing label smoothing~\cite{szegedy2016rethinking} level is beneficial.
\begin{table}[!t]
    \begin{center}
        \resizebox{.5\linewidth}{!}{
          \begin{tabular}{ l|c|c}
            \toprule
            & \multicolumn{2}{c}{EK100~\cite{ek100}}  \\
            & TSM & TSN  \\ 
            \midrule
            \textbf{DCR}                            &  {\bf 15.2} &  {\bf 14.5}   \\
            \midrule
            classification                          &    14.0     &  13.5         \\    
            \midrule
            $T_e$ = 1                               &    14.1     &  13.1         \\
            $T_e$ = 0                               &    14.6     &  13.9         \\
            linear $ T_e$                           &    15.0     &  14.2         \\
            exponential $T_e$                       &  {\bf 15.2} &  14.4         \\
            \midrule
            \textit{w.o.} $L_{rec}$                 &    14.5     &  13.8         \\
            \textit{w.o.} label smooth              &    14.0     &  13.3         \\
            \bottomrule
          \end{tabular}
        }
        \end{center}
    \vspace{-0.5cm}
    \caption{Ablation study of LSTM~\cite{lstm} version DCR.}
    \label{tab:abl_lstm}
\end{table}

\section{Additional Ablation Study}

We give more ablation study of DCR-LSTM in Tab.~\ref{tab:abl_lstm}. It has a little difference with the transformer reasoner since the order-aware pre-training is not used. Comparing with results in Tab.7 main text, we can conclude some properties between transformer and LSTM. First, LSTM is more robust with context as it doesn't collaspe when we set $T_e=1$ in training but $T_e=0$ in testing. Second, for LSTM, global easiness schedules (\eg exponential, linear) can achieve comparable performance with the local schedule in DCR. Third, transformer benefits more from feature level supervision but LSTM is less sensitive.

\begin{figure}
    \centering
    \includegraphics[width=0.7\linewidth]{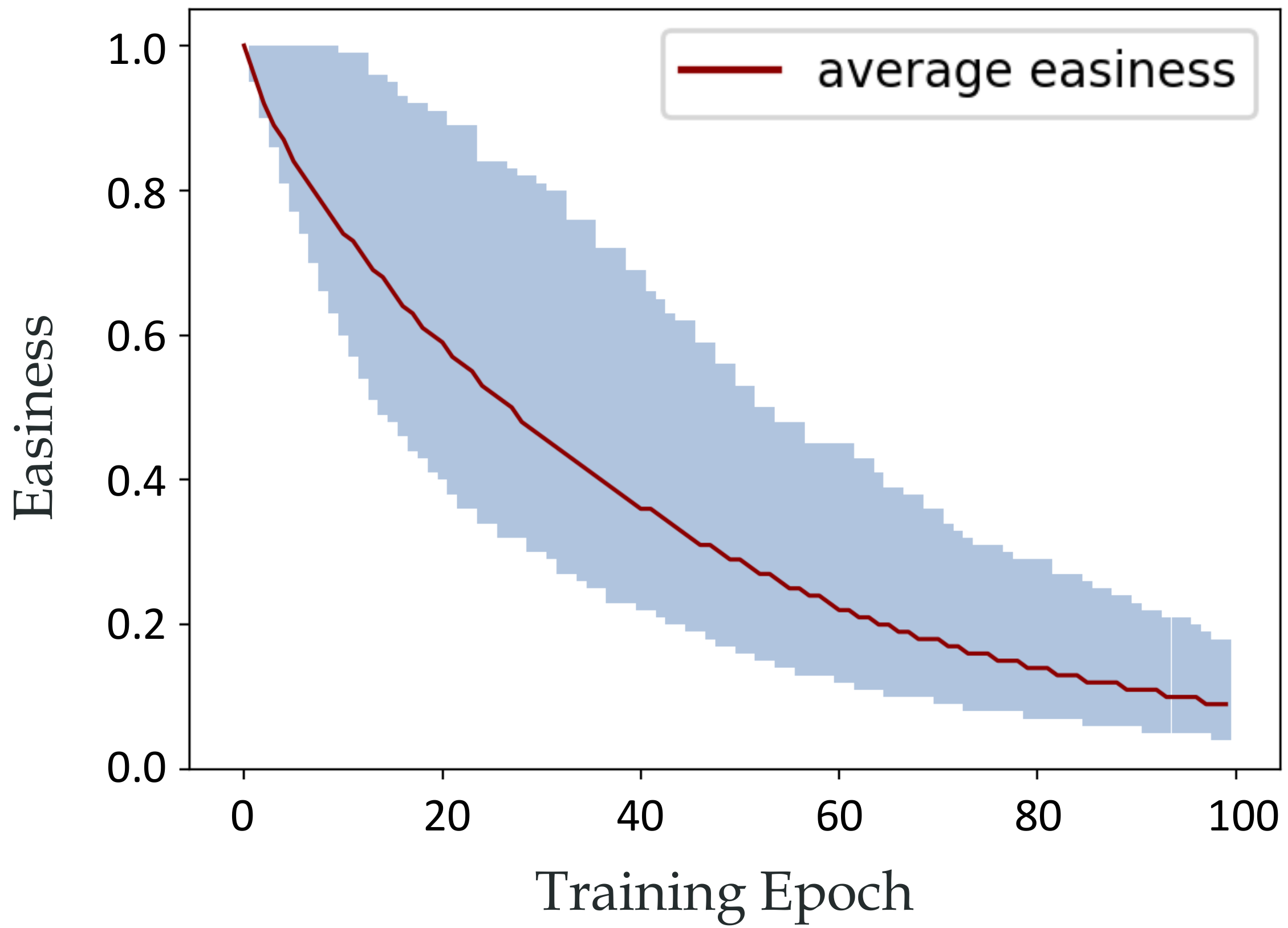}
    \caption{Easiness schedule for DCR training on EK100~\cite{ek100}.}
    \label{fig:easiness}
\end{figure}
\begin{table}[!t]
\begin{center}
\resizebox{0.45\linewidth}{!}{
  \begin{tabular}{lc}
    \toprule
                                    & Top-1 Acc        \\
    \midrule
    TCN~\cite{bai2018empirical}     &   19.1     \\
    RL~\cite{ma2016learning}        &   28.6     \\
    EL~\cite{jain2016recurrent}     &   27.5     \\
    RULSTM~\cite{rulstm}            &   30.9     \\
    DCR                             &   \textbf{33.0}   \\
    
    \bottomrule
  \end{tabular}
}
\end{center}
\vspace{-0.5cm}
\caption{Results of early action recognition with 50\% observation on EK55.}
\label{table:result_ear}
\end{table}
\section{Easiness Schedule}
Our instance-specific easiness schedule is visualize in Fig.~\ref{fig:easiness}. In each training epoch, we bound easiness range with minimal and maximal $T_e$ value among all instances and compute the average value as the red line in the figure. We can observe one fast easiness decreasing line as the bottom of the blue area while another slow easiness decreasing line as the upper bound. This indicates different difficulty in reasoning different video clips thus a finer-grained schedule is necessary with empirical supports.

\begin{figure*}[!t]
    \centering
    \includegraphics[width=\textwidth]{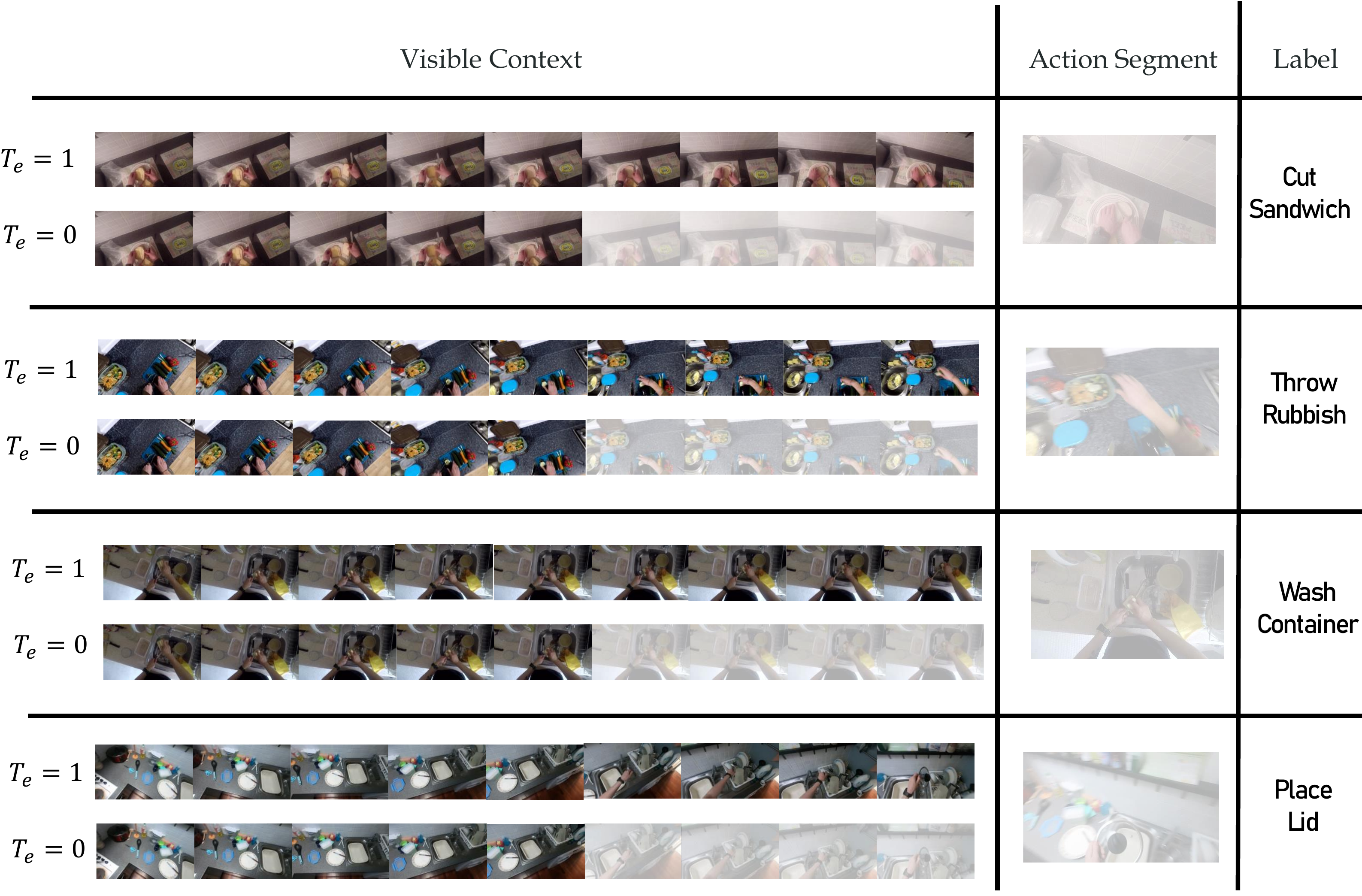}
    \caption{Cases of the anticipation task. Transparent frames are not visible.}
    \label{fig:case}
\end{figure*}

\section{Early Action Recognition}
As a general training strategy, DCR has the potential to support a wider range of temporal predictive applications, not limited to action anticipation. We take 50\% observation early action recognition as an example, \ie recognizing video action based on 50\% part video clips. Our method can have simple migration by modifying frame setting in Fig.3 and Eq.2 of main text. Detailed, we sample 40 frames from each action clip. First 50\% frames are constant
observation (blue in Fig.3 main text). Other 50\% frames have dynamic visibility (orange). One additional \texttt{[CLS]} is employed to predict label (yellow). Notably, we use the reconstruction quality of 1s future since last observation as easiness scheduling criterion in anticipation task in Eq.3 of main text. But it may not accessed in early action recognition scenario, we use the last frame reconstruction instead in a few exception cases. We conduct an experiment on EK55~\cite{ek55}, with the same setting following RULSTM~\cite{rulstm}. We train top transformer on RGB-TSN, FLOW-TSN, OBJ-FRCNN three backbones then late-fuse by 1:1:1 to obtain the final prediction. Results are listed in Tab.~\ref{table:result_ear}. We outperform baselines by a clear margin, validating the generalization ability of our method.

\section{Cases}
We show cases of anticipation task with easiness $T_e=0$ or $T_e=1$ in Fig.~\ref{fig:case}. Some key frames are not visible in the difficult mode of $T_e=0$.

\end{document}